# Multi-Objective Evolutionary approach for the Performance Improvement of Learners using Ensembling Feature selection and Discretization Technique on Medical data

Deepak Singh[1,*], Dilip Singh Sisodia[1], Pradeep Singh[1]
dsingh.phd2016.cs@nitrr.ac.in, dssisodia.cs@nitrr.ac.in, psingh.cs@nitrr.ac.in
[1]Department of Computer Science and Engineering, National Institute of Technology, Raipur, C.G, India

**Abstract**

Biomedical data is filled with continuous real values; these values in the feature set tend to create problems like underfitting, the curse of dimensionality and increase in misclassification rate because of higher variance. In response, pre-processing techniques on dataset minimizes the side effects and have shown success in maintaining the adequate accuracy. Feature selection and discretization are the two necessary preprocessing steps that were effectively employed to handle the data redundancies in the biomedical data. However, in the previous works, the absence of unified effort by integrating feature selection and discretization together in solving the data redundancy problem leads to the disjoint and fragmented field. This paper proposes a novel multi-objective based dimensionality reduction framework, which incorporates both discretization and feature reduction as an ensemble model for performing feature selection and discretization. Selection of optimal features and the categorization of discretized and non-discretized features from the feature subset is governed by the multi-objective genetic algorithm (NSGA-II). The two objective, minimizing the error rate during the feature selection and maximizing the information gain while discretization is considered as fitness criteria. The proposed model used wrapper-based feature selection algorithm to select the optimal features and categorized these selected features into two blocks namely discretized and non-discretized blocks. The feature belongs to the discretized block will participate in the binary discretization while the second block features will not be discretized and used in its original form. For the establishment and acceptability of the proposed ensemble model, the experiment is conducted on the fifteen medical datasets, and the metric such as accuracy, mean and standard deviation are computed for the performance evaluation of the classifiers. After an extensive experiment conducted on the dataset, it can be said that the proposed model improves the classification rate and outperform the base learner.

## 1. Introduction

Voluminous and complex nature of the biomedical data has created a great challenge in the analysis and design of *Insilico* models (1). Most of the *Insilico* model adopts the supervised learning for the diagnosis of the medical illness (2). Flowchart of the generalized supervised learning system is represented in Fig. 1. This block diagram shows the major operation involved during supervised learning. The primary requirement of the supervised learning problem is to have distinct domain-specific data with an appropriate number of features to overcome the complexity of storage and handling. Often, the biomedical datasets are burdened with higher numbers of redundancies including noisy, missing and irrelevant data. Knowledge extraction from these high dimensional data is a time consuming and labor intensive task. Thus, data processing has become a critical research topic in useful information extraction. In the Fig. 1 data reduction block aims to filter out the irrelevant data and minimizes the overhead storage complexity (3). The reduced data is termed as pre-processing data and will further participate in the training process of classifier/learner model.

Data dimensionality reduction (4), as a component of a data preprocessing step, is significant in numerous real-world applications. The dimensionality reduction has emerged since one of the considerable tasks in biomedical applications (5) and has already been effective in removing duplicates, increasing learning accuracy, and increasing decision-making procedures. Feature selection and discretization are the two-dimensionality reduction techniques that are employed to eliminate the data redundancies. A lot of literature proposal provides the utilization of feature selection techniques and discretization methods in biomedical data for achieving the dimensionality reduction which leads to improved classification rate. Feature selection has been a great active field of analysis for decades in data mining and has been widely applied to several biomedical fields including gene analysis (6), medical illness diagnosis (7), protein study (8) and drug discovery (9)(10). Discretization is another popular reduction methods which often convert continuous values into discrete values with fixed interval span (11). A substantial amount of literature is found in typically the utilization of discretization which often indicates the improved performance of learners (12).

Medical data comprises a mixed set of three major data types namely continuous, discrete and nominal. The range of continuous data for an attribute can be infinitely many whereas discrete and nominal are finite data. This wide range of continuous attribute values affects the classifier performance especially when the classifier follows the tree or rule-based approach (13). To cope with the difficulty faced by the learner, dimensionality reduction methods such as feature selection and discretization are employed that can lessen the

---

\* Correspondence author: Deepak Singh, Postal address: Department of Computer Science and Engineering, National Institute of Technology, G.E. Road Raipur, Chhattisgarh 492001, India. Tel: +91-9827916708

2effect. Feature selection removes the redundant data whose presence degrades the learner performance whereas discretization of data eliminates the inherent complexities associated with original attribute and instances. In the case of decision tree and rule-based learners, these dimensionality reductions help to form a smaller and easier tree and lesser rules compared to huge and complex tree shaped due to the higher number of feature and presence of the real-valued attribute in the input dataset. Feature selection technique is broadly categorized into three: filter, wrapper and embedded. Different techniques are adopted based on the model requirement and feature utility. In discretization, data partition is done through cut-point selection by applying different types of heuristics including Equal-interval-width (14), Equal-frequency-per-interval, minimal-class-entropy (15) and bunch based methods.

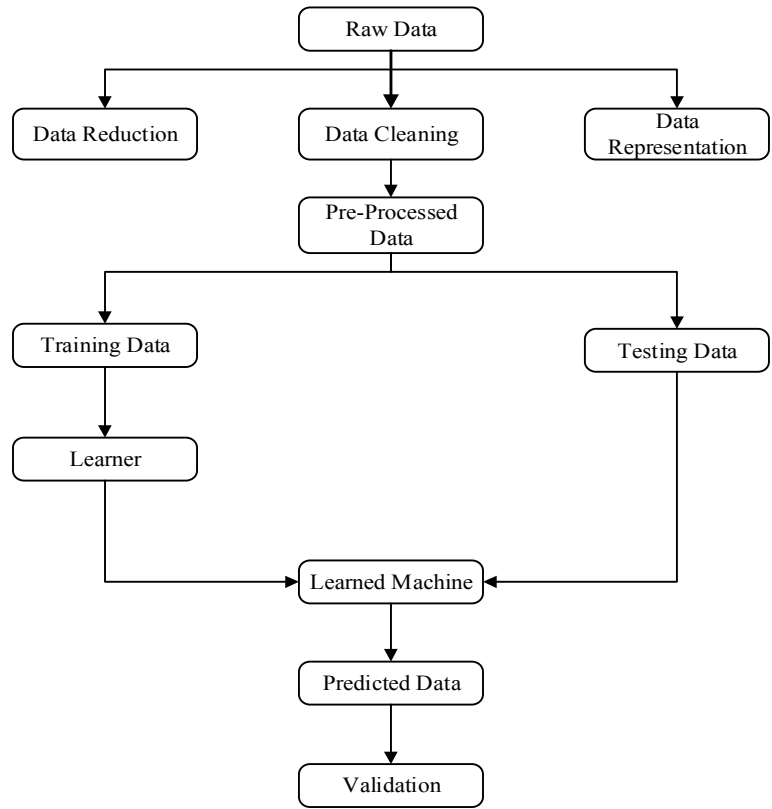

**Fig. 1: Generalized Supervised Learning Process**

Despite having distinctive characteristics and methodologies both feature selection and discretization suffers from high data affinity i:e different methods correspond to different notions of similarity and not every algorithm perform well on each dataset. Thus the selection of these algorithms is exhausting. Apart from this major drawback, there are other limitations like desired number of cut points as a variable, standard for your selection of optimum clusters and use of exhaustive search methods. Thereby to deal with these problems an integration of computational intelligence and machine learning tools is required that can render the extraction of information by examining the raw data. The objective behind the integration is to reduce the time consumed and minimize the classification error (16). Performance of machine learning algorithms highly relies on the data type and its distribution. With a given input raw data, it is very costly and challenging task to estimate the data distribution (17), so one can only choose by integrating intelligent techniques with machine learning for improving the performance even with redundant data.

Supervised learning in high dimensional complex data is inherently multiobjective. Many correlated objectives have often taken into consideration for the efficient learning. Single objective function, algorithm ignores the crucial factors that may affect the problem-solving. Therefore multi-objective algorithm is generally recommended that yield the set of the optimal solution by considering many criteria (18). In this context, multiobjective evolutionary algorithms can help in learning and provide a suitable working solution to complex high dimensional data handling. Since these algorithms are intelligent, can learn and help in adapting the dataset drift. There is the extensive usage of evolutionary computing found in the literature that advances the performance of supervised learning, however to the best of our knowledge, the existing literature does not support any method framed on evolutionary algorithm that addresses the issue of dimensionality reduction by incorporating feature selection and discretization simultaneously. This motivates us to explore whether a dimensionality reduction based on evolutionary computing framework can be applied to improve the prediction on various biomedical datasets.

Feature subset selection and cut-point selection for discretizing the attribute are search problems and can be achieved by the use of an Evolutionary algorithm (19)(20). The evolutionary algorithm in the past has been utilized successively for solving the search space problem (21)(22). Its application in the discipline of data processing has promising effects on the efficiency. Few attempts found on the evolutionary-based discretization whose objective is to increase the classification accuracy (20). In this paper, we extend and employ the robustness of multi-objective genetic algorithm NSGA-II (23) protocol for solving the optimal features and cut point selection in discretizing the feature. The novelty of the proposed model is to depute non-dominated sorting genetic algorithm for the removal of redundant data and simultaneously discretize the selective features obtained from the optimal feature set. The contribution of the proposed method can be summarized in the following key points:

- To the best of our knowledge, this is the first approach of solving the dimensionality reduction problem from the multi-objective criteria of minimizing the classification error and maximizing the information gain.
- The unified framework that uses feature selection and discretization simultaneously for the removal of domain level as well as attribute level redundancy.
- Information gain based heuristics are used for assessment of cut point selection, and the classification error rate is used for the feature selection evaluation.
- Utilization of Non-Dominated sorting genetic algorithm for achieving the desired multi-objective criteria. NSGA-II select the feature subset and generates the cut details for the selective features from the feature subset.
- Performance comparison of three popular baseline classifiers namely Decision tree, Naïve Bayes and Support vector machine on fifteen benchmark medical dataset. Additionally, to validate the exploration capability, Pareto analysis of NSGA-II is also performed.

The proposed algorithm generates only one cut point for selected optimal attribute inside the dataset, which often, in turn, results in the conversion of continuous attributes into the binary attribute. This binarization results in the shorter number of rules and trees that helps in the easy learning process with an additional benefit of reduced storage cost. Further, the rest of the paper is organized in the following manner; Literature work is described in section 2. The problem of features subset selection is defined in section 3.1, discretization of the features is presented in section 3.2. Evolutionary algorithm is discussed in section 3.3; the subsequent section 3.4 describes the proposed methodology, section 4 shows the feasibility study of the proposed system on the various analysis parameters and the last section 5 is making a concluding remark on the overall performance of the proposed model.

## 2. Related Work

Study and analysis in improving dimensionality reduction are common due to its higher applicability in the diversified problems. Feature selection (24) and discretization (25) are promising approaches to obtain this objective, which justifies its relationship to other procedures and problems. This section provides a brief review of topics closely related to dimensionality reduction from an assumptive and application point of view.

### 2.1 *Evolutionary Algorithm on Feature Selection*

The feature selection is accountable for selecting a subsection, subdivision or subgroup associated with feature space, which is often identified as search space problem. Fig. 2 shows a generalized feature selection process bearing the elementary five component, where subset evaluation is achieved by utilizing a fitness function to measure the goodness/quality of the selected features. Generally, Feature selection algorithms are classified into two categories: filter based (26) and wrapper-based (27). The difference between the approaches is the use of classifier or learning algorithm as fitness evaluation criteria in the wrapper based algorithms. However, there is one more approach existed other than the two and termed as embedded which is a hybrid combination of the filter and wrapper techniques (28). Feature selection algorithms have the primary objective of maximizing the classification rate with a minimum number of features. There are a lot of literature available on the feature selection problems and can be seen in (29), here to simplify the organization of paper we follow the solutions being achieved from the evolutionary algorithm on wrapper approaches.

Many different evolutionary algorithms and its variants have been proposed to improve the performance, which primarily works on the representation, fitness design and search mechanisms for extracting an optimal number of features. In an early work, Genetic algorithm is employed to feature selection problem by analyzing the influence of the chromosome size, crossover, and mutation operators (30). In a variant of GA, a cooperative co-evolutionary algorithm for feature selection based on three populations has been proposed in (31), where the first population is targeted on feature selection, the second population is for instance selection, and the third population considering both feature selection and instance selection. A cluster-based feature selection technique is proposed in (32) where GA is used to optimize the cluster center for obtaining optimal feature clusters. Each feature in the cluster is ranked according to the distance from the cluster center. Top-ranked captured from each cluster is represented for feature selection. Modification in the chromosome encoding named as the bio-encoding scheme is proposed in (33), where each chromosome is represented by a pair of strings. The first vector is binary-encoded to show the selection of features, and the second is a vector of real number indicated the weights of features. Weights of the features are used by AdaBoost ensemble algorithm. There are few works that emphasize on the design of fitness function for the optimal feature subset selection. Classification accuracy and the number of features (34), accuracy and the cost of an ANN (35), the area under curve of the receiver operating characteristic of an NB classifier (36) were utilized as a single fitness function in a GA.

Feature selection involves two primary objectives, maximizing the classification accuracy and minimize the number of features. Both the objectives are contradictory. Thus, feature selection can be formulated as a multi-objective problem to find a set of trade-off solutions between these two objectives. There are few feature selection methods that engage multi-objective genetic algorithm. In (37) NSGA-II is utilized to select the local feature subsets of various sizes in the churn prediction problem, overall accuracy, true churn accuracy, and true non-churn accuracy are the three fitness functions considered here. An SVM based wrapper feature selection for diagnosing cancer from miRNA sequence is proposed in (38), the three objective of sensitivity, specificity, and number of features are simultaneously being optimized by NSGA-II. Mostly non-dominated sorting algorithm (NSGA-II) or its variations is elected for achieving the multi-objectivity in the feature selection problem. NSGA-II remains the best choice for the feature subset selection due to its ability to explore the non-dominated solutions efficiently and satisfactory past performance on various applications.

## 2.2 Background on Discretization

Discretization process transforms real numerical attributes into a discrete or nominal attribute. Enumerating the various techniques a taxonomy of discretization methods are present here. Different heuristic approaches have been proposed for discretization, for example, Chi-square, it is a variable used to look for the similarity between intervals. This statistics is employed in Chi-Merge, for partitioning the continuous characteristics (39). This technique does not consider the class information for cut point selection. CAIM (class-attribute inter-dependence maximization) (40) is a technique based on class-attribute inter-dependence heuristics which finds the minimum number of cut points with an objective of maximizing class-attribute dependence heuristics. In (41), the author addresses the utilization of entropy-based heuristics for discretizing the range of constant attribute into numeric interims. Principal component analysis (PCA) based discretization algorithm proposed in (42) which generate the interval by exploiting the particular correlation structure between the attributes. A parallel distance-based measure was demonstrated in (43) which has an adverse effect on time complexity. AMEVA (44) is a discretization algorithm which generates a potentially minimal number of discrete intervals by maximizing a new contingency coefficient based on Chi-square statistics. The benefit of the AMEVA is that it does not require the user to explicitly mention the number of required intervals.

MODL (45) discretization method is based on Bayesian approach and proposed Bayes optimal examination criterion for cut-point selection where a new super-linear optimization formula manages in finding the near-optimal intervals. The two efficient unsupervised, proportional discretization and stuck frequency discretization methods are proposed in (46) that can effectively manage discretization bias in addition to variance. The concept of greedy method and class attribute contingency coefficient is utilized in (11) with a motive of improving the data quality regarding discretized data. Evolutionary computation based algorithms are also noticed for the cut-point selection problems. In this context, subset selection of cut points with the utilization of evolutionary algorithm for identifying the best splitting criterion using a wrapper fitness function is proposed in EMD algorithm (47). Multi-objective evolutionary protocol approach with the goal of minimizing classification error (the first objective function) and a number of cut details (the second objective function) are simultaneously performed in (48). The potential of evolutionary algorithms (EA) for multi-objective discretization is still needed to be investigated since discretization is a complex task and requires specifically designed multi-objective EAs to search for the non-dominated solutions.

## 3. Dimensionality Reduction by Ensemble Approach

This section describes the proposed dimensionality reduction in biomedical data for the sake of easy adaptation and performance improvement. We further divide this section into four subsections: a formal introduction on feature selection is described in the first subsection, discretization in the second, Evolutionary algorithm in the third subsection and proposed ensemble model in the fourth subsection.

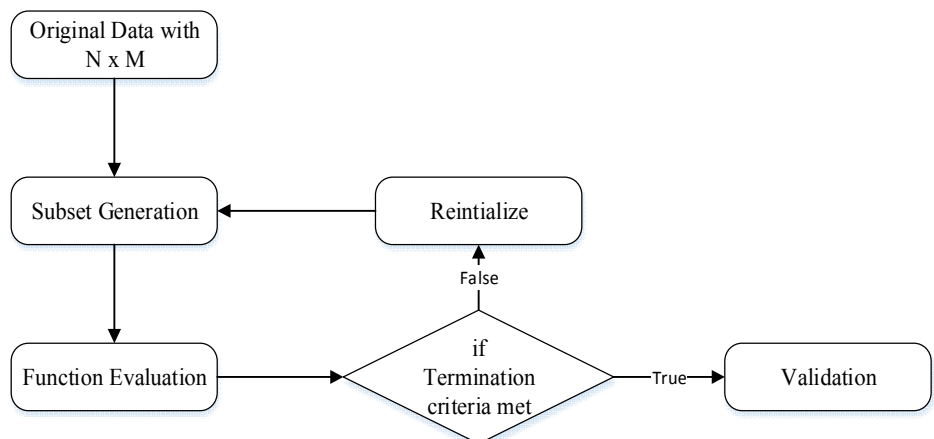

**Fig. 2: Generalized Feature Selection Process**

## 3.1. Feature selection

Feature section problem can be well understood by the mathematical model of maximizing the classification evaluation functions while keeping the minimum number of a feature from the feature set (49). To define formally, Let $f(c)$ is classification evaluation function with a characteristic of maxima and its values $f(c) \epsilon\ [0,1]$ where 1 yields the ideal best performance. Data of size $m \times n$ matrix denotes the $m$ number of instances and $n$ number of features. The objective of feature selection is to obtain the data of size $m \times n'$ such that $n' < n$ and the value of $f(c)$ being the maximum. There are no constraints put on the data type of feature, it can be discrete, real or nominal. There is the probability of the presence of more than one subset solution whose function value $f(c)$ being optimal or same.

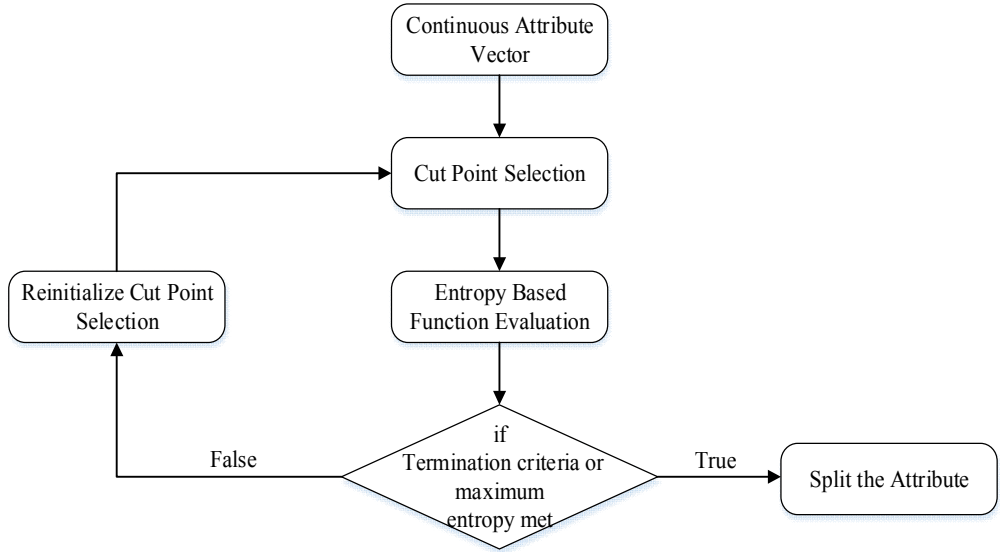

Fig. 3: Generalized Discretization Process

## 3.2. Discretization

The problem of cut-point selection in discretization is one of the partitioning criteria for converting the continuous feature values into the discrete interval (50). It has been proven that the process of discretization improves the performance of classifier (51). It reduces a large number of real-valued into the discrete space by combining few of the values together. For an optimal discretization method, the choice of the attribute relies on class dependency relationship. Class-entropy is the attribute-class relationship commonly used as a criterion for the discretization. A generalized process of discretization is illustrated in Fig.3.

To define discretization mathematically, Let $A$ be a vector of the numeric attribute, and assuming the domain of $A$ be in the interval of $[lb, ub]$. For any instance i from the dataset $D$, also have one of the values from the attribute $A$, For example, considering the dataset $D$ given for the classification of illness amongst the various person. This dataset has various attributes where one of the attributes $A$ stands for the weight of the persons if at index $i$ of the attribute $A$ has value 69.32 kg then this means that $i^{th}$ person weight is 69.32 kg and will be considered for the prediction of some illness. Mostly in classification problem, several instances of numeric and nominal attributes form a feature matrix, there is also class label associated with each of the instances. $C$ is the vector of the class label of nominal data type following the rule $A(i) \rightarrow C(i)$.

Let $P_a$ be the partition on attribute $A$ and it is defined as the following set of $k$ subintervals:

$$P_a = \{[I_1, I_2, I_3, \ldots, I_k]\} \quad k >= 2$$

where $I_1 = [d_0, d_1], I_j = [d_{j-1}, d_j], I_k = [d_{k-1}, d_k]$

$d_0 = lb$, $d_{j-1} < d_j$ and $d_k = ub$, $I_j$ is the interval within the range $d_{j-1\ldots j}$. These ranges are the interval bound values of the continuous attribute which will be used for the conversion into the discrete form.

$$\textbf{\textit{if }} d_{j-1} \leq A(i) < d_j \textbf{\textit{ then }} A(i) = I_j$$

Binary discretization is the process where the partition ranges into two intervals. A threshold value $th$ is determined which partitioned the data into two by

$$\textbf{\textit{if }} A(i) <= th \textbf{\textit{ then }} A'(i) = 1$$

$$\textbf{\textit{else }} A'(i) = 0$$

Information gain is the class-based heuristic approach used in the decision tree for splitting the tree branches. This heuristics can also be used as a discretized measure. To understand, let's suppose that there are $n_c$ numbers of classes in the vector $C$ and a partition $P_a$ produces two subsets $S_1$ and $S_2$. Then the class entropy for each subset is computed by

$$Ent(S_i) = - \sum_{k=0}^{n_c} P(C_k, S_i) \log P(C_k, S_i) \quad (1)$$

Entropy is the measure of information required in bits to denote the class $C$. The two partitioned subset $S_1$ and $S_2$ can be evaluated by the weighted average of the individual class entropy.

$$Ent(S, A) = \frac{|S_1|}{|S|} Ent(S_1) + \frac{|S_2|}{|S|} Ent(S_2) \quad (2)$$

$$Gain = Ent_{cl} - Ent(S, A) \quad (3)$$

$Ent_{cl}$ is the entropy of class label $C$. The gain value is the difference between the class entropy and the weighted partition entropy. For the best partition $P_a$, this gain value should be maxima and can be achieved through the iterative procedure.

In the proposed model, selected features $D_f$ is being generated by feature selection algorithm, and $D_{fd}$ is the selected continuous feature from the $D_f$ i.e. $D_{fd} \epsilon D_f$ for discretizing the attribute. For each $D_{fd}$ the choice of a cut-point $f_t$ from the range of the corresponding attribute is made on the basis of maximum gain Eq.(2). In order to pick the optimal cut point, the need of the system is to evaluate each split point but that is only possible through an exhaustive search.

### 3.3. Evolutionary algorithm

Darwin's theory of survival of the fittest is the origin of Evolutionary algorithm. The algorithm mimics the natural process of evolution. There are many algorithms lie in this category out of them the most widely used is Genetic Algorithm (52). It is an evolutionary-based algorithm where the set of chromosomes formed by the genes are trying to produce a better solution with the help of genetic operators in an iterative way. Some random solutions are formulated as chromosomes and genes at an initialization step, and the process of crossover and mutation takes place in a defined way to improve the potential solution.

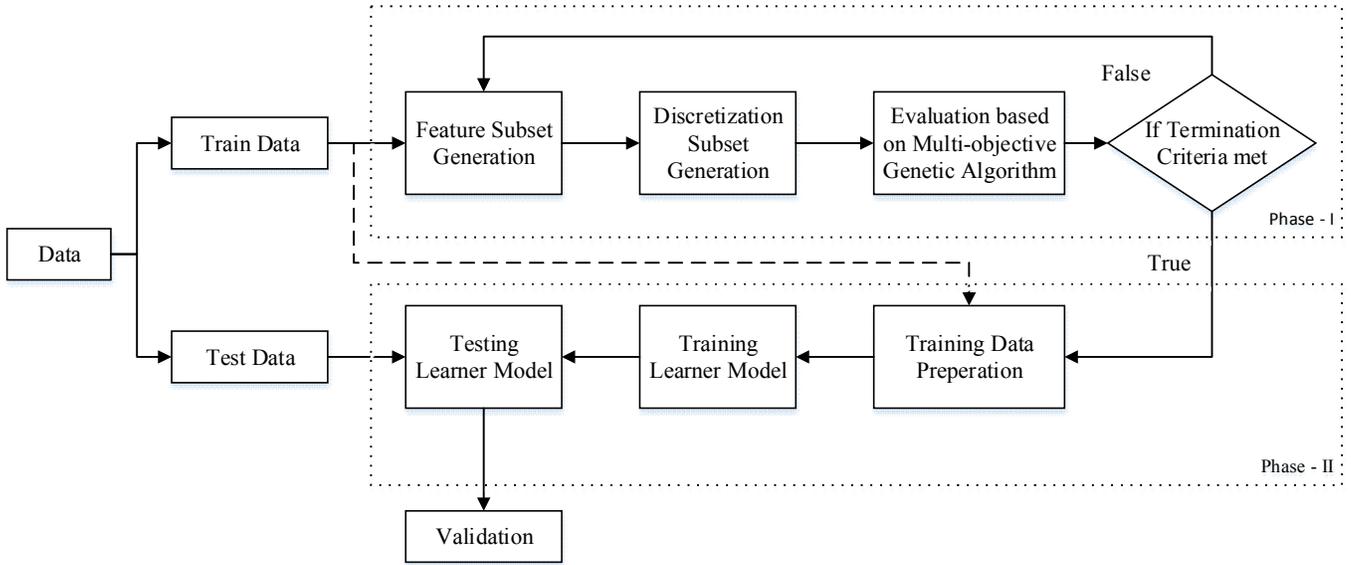

**Fig. 4: Work Flow of the Proposed Ensemble Model**

With the motivation to produce multiple trade-off solutions between reduced features and performance for classification problem, we have used an idea of multi-objective optimization. In multi-objective optimization, multiple solutions called *Pareto solutions* or *non-dominated solutions* are produced by considering more than one objective simultaneously. A solution is called Pareto solution, if it is better in at least one objective as compared to other Pareto solutions. There are various variant present and proposed, here we employed a controlled elitist genetic algorithm (a variant of NSGA-II) (23) due to its high success rate in achieving the dimensionality reduction.

### 3.4. Proposed Ensemble of Feature selection and Discretization

This section describes the evolutionary-based unified dimensionality reduction framework where feature selection and discretization perform co-operatively; Non-dominated sorting genetic algorithm is applied for acquiring the desired model. The two objectives of maxima information gain Eq.(4) and minima classification error Eq.(5) is considered respectively. The workflow diagram of the proposed unified model is given in Fig. 4.

NSGA-II is a population-based search approach where each individual of the population is termed as a *chromosome*. The chromosome of the algorithm is a combination of hundreds and thousands of genes, and the size of a gene can be varied. In this coding of individual representation, the numbers of genes are fixed to the number of features in the dataset. Each gene will be of three dimensions encoded as the hybrid vector of binary and real values. The first dimension of the gene is a binary value indicating the presence or absence of the feature. The second dimension of the gene presents the binary value for the attribute to be binarily discretized or not, but this binarization is dependent on the feature selected index. The third dimension of the gene gives the cut-point value for the binary conversion.

$$maximize\ fit(1) = \sum_{j=1}^{n} Gain_j \qquad (4)$$

$$minimize\ fit(2) = Err \qquad (5)$$

For the initialization cut-value vector of the chromosomes or individual, the minimum and maximum domain value of each attribute is calculated; these minimum and maximum values are the boundary points and are considered as lower bound ($lb$) and upper bound ($ub$). The random initialization of the initial population uses these $lb$ and $ub$ vectors. In this way, the chromosomes consist of random value lying in the domain $[lb, ub]$ of attributes. Not all the attributes required discretization thus in addition to it, each gene has associated binary value that represents the presence and absence of corresponding attribute in discretization. This will produce the optimum sequence of discretized attributes and also save the computational effort required for each attributes.

To simplify, we consider an example on Iris dataset shown in Fig. 5 that consists of 50 samples from each of the three species of iris flower (Setosa, Virginia, and Versicolor). There are four features with a total instance of 150 with an objective of classifying the species with these features.

| S.No. | S_len. | S_wid. | P_len. | P_wid. | Flower |
|---|---|---|---|---|---|
| 1 | 5.1 | 3.5 | 1.4 | 0.2 | Setosa |
| . | ... | ... | ... | ... | ... |
| 51 | 7 | 3.2 | 4.7 | 1.4 | Versicolor |
| . | ... | ... | ... | ... | ... |
| 150 | 5.8 | 2.7 | 5.1 | 1.9 | Virginica |

**Fig. 5: IRIS dataset example**

The structure of an individual in genetic algorithm for finding out the optimal solution is shown in Fig. 6; this single chromosome consists of length $n$ genes, where $n$ is the number of features. In iris data, the value of $n$ is 4, and each gene will be of three dimensions representing the first dimension for feature selection second for discretization and third to be assigned for the cut point value.

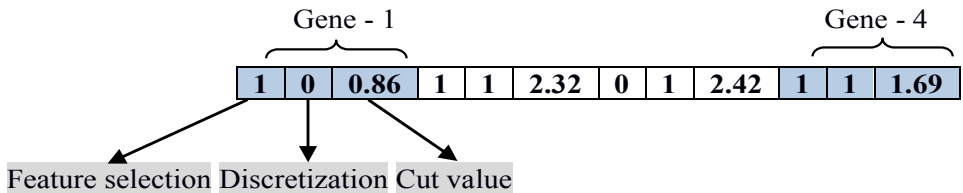

**Fig. 6: Structure of Chromosome**

From the given instance (Fig. 6) of chromosome, individual gene 1 selects the first feature but won't be considered for discretization, gene 2 and 4 both are selected features and also considered for discretization along with indicated cut points whereas gene 3 does not select the feature for evaluation. Algorithm 1 decodes the chromosome structure.

| Algorithm – 1 Chromosome Decoding |
|---|
| **Procedure Gene_Decode ($gene$)** |
| 1. $if\ gene_i(1) = 1$ |
| 2. $\quad feature\ is\ present$ |
| 3. $\quad if\ gene_i(2) = 1$ |
| 4. $\quad\quad feature\ considered\ for\ discretization$ |
| 5. $\quad\quad use\ the\ gene_i(3)\ value\ for\ the\ binarization$ |
| 6. $\quad end$ |
| 7. **else** |
| 8. $\quad feature\ is\ absent$ |
| 9. **end** |

| S. No. | S_len | S_wid(D) | | P_wid(d) | | Flower |
|---|---|---|---|---|---|---|
| 1 | 5.1 | 3.5 > 2.32 | = 1 | 0.2 < 1.69 | = 0 | Setosa |
| ... | ... | ... | ... | ... | | ... |
| 51 | 7 | 3.2 > 2.32 | = 1 | 1.4 < 1.69 | = 0 | Versicolor |
| ... | ... | ... | | ... | | ... |
| 150 | 5.8 | 2.7 < 2.32 | = 0 | 1.9 > 1.69 | = 1 | Verginica |

**Fig. 7: Decoding of chromosome**

Wrapper-based feature selection through evolutionary algorithm is done previously, in this study we followed (53) to obtain the optimal features. For the cut point evaluation, the objective of maxima gain is initiated by calculating the class entropy. Few of the dataset has more than two class label thus entropy can be calculated by using Eq. (1), In the iris dataset, the total instances are 150 where out of 150 data 50 instances are 'Setosa,' 50 are 'Versicolor' and the last 50 are 'Virginia', so class entropy calculated as:

$$Ent_{cl} = -((50/150) * log(50/150) + (50/150) * log(50/150) + (50/150) * log(50/150))$$

$$Ent_{cl} = 1.5850$$

**Algorithm – 2 Multi-Objective Evolutionary based Dimension Reduction (MOEDR)**

**Procedure MOEDR ($X, npop, ntime, bound$)**

Input:
    $X$                                                                                      // Dataset of size m instances and n attributes
    $npop$                                                                 // Population size
    $ntime$                                                     // Maximum Iterations
    $bound$                                        // Lower and upper bound for Initialization
1. $pop \leftarrow Initialize\_Pop\_Random\ (npop, range)$
2. $fit\_np \leftarrow Compute\_Fitness(pop, X)$                 //Calling procedure **CEF** and **CAF** for fitness
3. $front \leftarrow Fast\_Non\_Domination\_Sort\ (pop, fit\_np)$           // Front Generation
4. $rank \leftarrow Assign\_Rank(front, pop)$               // Each individual according to their fronts
5. $dist \leftarrow Crowding\_Distance(pop, rank)$
6. $set\ t \leftarrow 0$
7. **while** $t < ntime$ **or** termination criteria met **do**
8.     $qpop(t) \leftarrow Genetic\_Reproduction(\ pop(t), fit\_np)$     // Selection, crossover, mutation
9.     $fit\_nq \leftarrow Compute\_Fitness(qpop(t), X)$
10.     $R(t) \leftarrow P(t) \cup Q(t)$           //Merge the newly generated solution with the original solutions
11.     $front(t) \leftarrow Fast\_Non\_Domination\_Sort\ (R(t), fit\_nq)$
12.     $rank(t) \leftarrow Assign\_Rank(front(t), R(t))$
13.     $pop(t+1) \leftarrow [\ ]; i \leftarrow 1$                    // Initialize next population
14.     **while** $|pop(t+1) + front(t)| < npop$
15.         $fron_i \leftarrow Crowding\_Distance(R(t), rank(t))$
16.         $pop(t+1) \leftarrow pop(t+1)\ U\ fron_i$
17.         $i \leftarrow i + 1$
18.     **endwhile**
19.     *Sort and add top **npop** element from the front in $pop(t+1)$*
20.     $t \leftarrow t + 1$
21. **endwhile**
22. *Report the final members of pop as final solution*

According to the chromosome structure shown in Fig. 6, the second attribute is selected for discretization with cut value 2.32. Therefore, attribute with values greater than 2.32 will split into one partition (replace values with 1) and value lesser than 2.32 will lead to the other partition (replace values with 0). This will result in two new subsets $S1$ and $S2$. Fig. 7 shows the resultant transformed table obtained when individual chromosome decoded. These transformed data will be used for wrapper feature assessment and entropy based cut point evaluation. The entropy of obtained subset $S1$ and $S2$ is required to evaluate the cut-point, therefore entropy of subsets are:

$$Ent_{S1} = -((49/142) * log(49/142) + (44/142) * log(44/142) + (49/142) * log(49/142))$$

$$Ent_{S1} = 1.5832$$

$$Ent_{S2} = ((1/8) * log(1/8) + (6/8) * log(6/8) + (1/8) * log(1/8))$$

$$Ent_{S2} = 1.0613$$

The assessment of cut-point two partitioned subset $S1$ and $S2$ is achieved by the weighted average of the individual splitting class entropy which is calculated as

$$Ent_S = (142 / 150) * 1.5832 + (8 / 150) * 1.0613$$

$$Ent_S = 1.5554$$

The gain value of the partitioning point of the selected attribute is computed by subtracting the weighted splitting entropy with class label entropy.

$$gain_{i,j} = 1.5850 - 1.554 = 0.0296$$

Finally the fitness value is obtained by adding the gain value of each selected discretized attribute by utilizing Eq. 4.

**Algorithm – 3 Discretization Fitness Evaluation**
**Procedure Calculate_Entropy_Fitness ( $X, Y, pop_i$ )**
Input: $X$ : Dataset      $Y$ : Label      $pop_i$ : $ith$ $chromosome$ $form$ $the$ $population$ $pool$
1. **while** $length$ $of$ $the$ $individual$ $chromosome$ **do**
2.      $split$ $the$ $attribute$ $by$ $the$ $cut$ $point$ $value$
3.      $Transfrom$ $attribute$ $into$ $binary$ $S1$ $and$ $S2$
4.      $Calculate$ $the$ $entropy$ $of$ $the$ $S1$ $and$ $S2$ $by$ $calling$ **procedure Calculate_Entropy**
5.      $Compute$ $Si$ $the$ $weighted$ $average$ $entropy$ $of$ $S1$ $and$ $S2$
6.      $Gain_{i,j} \leftarrow Ent_{cl} - S_j$
7. **endwhile**
8. $fit(i) \leftarrow \sum_{j=1}^{n} Gain_{i,j}$
9. **return** $fit$

The multi-objective genetic algorithm starts with the random initialization of chromosomes further each chromosome is evaluated by the fitness function. Here we, consider two fitness measure: maxima summation of gain value and minima classification error rate. It then evolves initial population by selection, crossover and mutation operator. For the process of reproduction the selection of the fittest parent is the essential requirement, so to meet the criteria, tournament selection technique is used. In the tournament scheme, the probability of selecting the fittest parent is directly associated with the fitness value. The fittest parent then performs crossover operation to generate the new solutions.

**Algorithm – 4 Classification Error Rate for Wrapper based Feature Selection**
**Procedure Calculate_Error ($X, Y, pop_i$)**
Input: $L_{m \times 1}$ : $label$      $D_{m \times n}$: $Dataset$
1. $Selected$ $Features$ $Indices$ $S_{ind}$
2. $D_{sel} \leftarrow D\,[:, S_{ind}\,]$      // Selected feature Dataset
3. $Applying$ $K\text{-}fold$ $cross$ $Validation$      // 10-Fold Validation
4. **for each** $fold$ $i$ **do**
5.      $Train$ $the$ $Learner$ $with$ $the$ $selected$ $training$ $data$ $and$ $class$ $label$
6.      $Test$ $the$ $learner$ $with$ $the$ $testing$ $data$
7.      $Perform$ $the$ $validation$ $(accuracy = ACi)$
8.      $S \leftarrow S + AC_i$
9. **endfor**
10. $Compute$ $the$ $average$ $accuracy$ $by$ $AVG_{ac} = mean(S)$
11. **return** $AVG_{ac}$

Post reproduction, Gaussian mutation is performed. Since elitism is introduced by comparing current population with previously found best-nondominated solutions, the procedure is different after the initial generation. Set of operation (front calculation by non-domination sorting, rank computation and then elimination of low-rank population from the pool) performs in an iterative way to evolve into optimal ensembles (23).

**Algorithm – 5 Entropy Calculation for the Partitions**
**Procedure Calculate_Entropy ($L$ )**
Input: $L$ : the class label
1. **for each** $unique$ $label$ $in$ $class$ $label$ $L$ **do**
2.      $Find$ $the$ $length$ $of$ $indices$ $of$ $label$ $i$
3.      $Calculate$ $the$ $entropy$ $of$ $label$ $i$ $by$ $using$ $Eq.\,(1)$
4.      $S = S + entropy$ $of$ $label$ $i$
5. **endfor**
6. **return** $S$

The algorithm for the proposed system is presented by five pseudo codes named Genetic Decode (Gene_Decode), Multiobjective Evolutionary based Dimension Reduction (MOEDR), Calculate Entropy Fitness, Calculate Error Fitness and Calculate Entropy (CE). These are designed to achieve the objective of dimensionality reduction. Algorithm – 2 is the main module that makes the feature selection, and discretization with maxima gain achievable using evolutionary cycle.

## 4. Experiment setup and Result

The objective of this work is to explore the importance of dimensionality reduction in biomedical data accomplished through a joint effort of feature selection and discretization. Therefore, accordingly experimental study is partitioned into three parts. In the first part, we listed the experimental setup and details of medical benchmark dataset. The second part carries the performance measurement of the proposed model on three baseline classifiers including Support vector machine, Naïve Bayes, and Decision tree. The third part presents the characteristics of the multi-objective algorithm (NSGA-II) with the analysis of non-dominated solutions also called Pareto-optimal front in dimensionality reduction framework.

### 4.1. Experimental Setup

For the experiment, a total of 15 standard biomedical benchmark datasets are considered. This dataset has been obtained from the publically open UCI machine learning repository (54). Their details are listed in Table 1which includes dataset name in the first column; number of attributes or feature present in the dataset represented in the second column, the third column presents the number of classes exists in the dataset, numbers of instances are shown in the fourth column, and the last column represents the number of continuous features present in the dataset. On observation, we draw the following details from the table

- Coil2000 is the largest dataset include more than 9000 instances of the two classes.
- Ratio of continuous attributes with respect to the total number of features in the dataset indicates the need for discretization.
- Majority of the referred dataset belongs to the binary class (9 out of 15).
- Presence of class imbalance ratio in few datasets (Abalone, E-coli, Pima and new thyroid).

**Table 1: Characteristics of the benchmark dataset**

| Dataset | Attribute | Class | Instances | Continuous Attribute |
|---|---|---|---|---|
| Abalone | 8 | 28 | 4174 | 7 |
| Appendicitis | 7 | 2 | 106 | 7 |
| Cleveland | 13 | 5 | 297 | 13 |
| Coil2000 | 85 | 2 | 9822 | 85 |
| Dermatology | 34 | 6 | 358 | 34 |
| Ecoli | 7 | 8 | 336 | 7 |
| Heart | 13 | 2 | 270 | 13 |
| Hepatitis | 19 | 2 | 80 | 19 |
| New thyroid | 5 | 3 | 215 | 5 |
| Pima | 8 | 2 | 768 | 8 |
| Saheart | 9 | 2 | 462 | 8 |
| Spectfheart | 44 | 2 | 267 | 44 |
| Thyroid | 21 | 3 | 7200 | 21 |
| Wdbc | 30 | 2 | 569 | 30 |
| Wisconsin | 9 | 2 | 683 | 9 |

All the experiments reported in this section were performed using an Intel core i7 3.4 GHz processor as CPU, 8 GB of main memory, and running windows 7. The proposed model is implemented in Matlab development environment. We reported the observed accuracies an average of 10 runs on 10-fold validation.

**Table 2: Parameter values**

| Method | Parameters |
|---|---|
| C4.5 | Pruned tree, confidence = 0.25, 2 items per leaf |
| Naïve Bayes | Distribution = 'Gaussian' |
| SVM | Kernel = 'Linear', method = 'QP' |
| NSGA-II | Crossover Type = scattered<br>Population = 50, Generations = 1000, Function Tolerance = 1e-3<br>Crossover rate = 0.8, max stall time = 600 |

411Learners are parameter sensitive; their performance is highly dependent on the input parameters. We reported the parameters of learner and NSGA-II in Table 2. Rows of the table are categorized into Base classifiers and multi-objective genetic algorithm Parameters. These values were set during the experimentation and result evaluation. Decision tree C4.5 with 10-fold cross validation is used as a function evaluator in the wrapper-based feature selection.

*4.2. Results and Discussion*

The experimental result on the standard benchmark dataset is discussed in this section. The proposed model minimizes the classification error rate on selected features and discretizing some subsets of selected features to overcome the difficulty in learning complex continuous attributes. This multi-objective problem is solved by NSGA-II in this paper. The multi-objective genetic algorithm produces a set of a solution called Pareto front non dominated solutions whereas single objective generates a single optimal solution. The choice of only optimal solution from the nondominated solutions is difficult because of a large number of solutions and absence of such solution which strictly dominates all solutions in all objectives. Therefore a solution with higher accuracy rate amongst the front is considered. The choice of higher accuracy rate is due to the requirement of high classification performance.

Three well-established baseline classifiers Decision tree C4.5, Naïve Bayes (NB) and Support vector machine (SVM) are selected for the performance evaluation of the proposed framework. The reason for the selection of these classifiers especially (C4.5, NB) are because these algorithms are only designed to work for nominal feature spaces, therefore even if the features are not discretized as a part of pre-processing work, the algorithm itself does some sort of discretization during the learning process.

Table 3: Naïve Bayes performance comparison on Accuracy measure

| Data Name | Features | Selected Features | Discretized | No. of fronts | Accuracy | Std. | Proposed Accuracy | Std. |
|---|---|---|---|---|---|---|---|---|
| Abalone | 8 | 6.19 | 4.86 | 15.29 | 0.254 | 0.020 | **0.264** | 0.019 |
| Appendicitis | 7 | 4.09 | 1.58 | 13.97 | 0.848 | 0.094 | **0.877** | 0.089 |
| Cleveland | 13 | 5.87 | 2.39 | 20.72 | 0.559 | 0.056 | **0.590** | 0.056 |
| Coil2000 | 85 | 37.09 | 31.21 | 23.95 | 0.914 | 0.004 | **0.930** | 0.014 |
| Dermatology | 34 | 17.37 | 8.04 | 14.90 | 0.922 | 0.049 | 0.910 | 0.037 |
| Ecoli | 7 | 5.90 | 2.66 | 22.18 | 0.830 | 0.048 | 0.818 | 0.050 |
| Heart | 13 | 6.72 | 4.96 | 14.31 | 0.811 | 0.085 | **0.837** | 0.080 |
| Hepatitis | 19 | 11.45 | 3.15 | 15.68 | 0.875 | 0.083 | 0.863 | 0.124 |
| Newthyroid | 5 | 4.04 | 2.07 | 11.05 | 0.967 | 0.032 | **0.981** | 0.033 |
| Pima | 8 | 4.04 | 2.32 | 30.97 | 0.734 | 0.054 | **0.780** | 0.027 |
| Saheart | 9 | 6.84 | 4.24 | 16.56 | 0.671 | 0.059 | **0.721** | 0.061 |
| Spectfheart | 44 | 20.03 | 7.26 | 13.62 | 0.734 | 0.099 | **0.809** | 0.071 |
| Thyroid | 21 | 15.04 | 13.61 | 18.23 | 0.996 | 0.003 | **0.998** | 0.003 |
| Wdbc | 30 | 15.51 | 7.34 | 26.20 | 0.942 | 0.039 | **0.949** | 0.034 |
| Wisconsin | 9 | 3.20 | 1.83 | 22.81 | 0.966 | 0.021 | **0.974** | 0.027 |

Table 3, 4 and 5 list the readings that were obtained during the experimentation performed on the standard medical dataset. These tables carries nine columns where the first column depicts the dataset name, the second column shows the number of feature in the dataset, the third column lists the average number of selected features by the proposed model, the fourth column informs the average number of discretized features amongst the selected features, average number of Pareto fronts given by NSGA-II is depicted in the fifth column, the consecutive two columns are accuracy and standard deviation of 10 runs on the baseline classifiers and the last two successive columns represent the accuracy and standard deviation produced by the proposed model. The best performance values are highlighted and represented in bold letters.

Reported results of Naïve Bayes classifier on the medical data are listed in Table 3. We performed the 10 runs of k-fold validation on each dataset with and without proposed unified ensemble dimension reduction model on the Naïve Bayes classifier. Based on the experiment, we observe the following facts

- Proposed model outperform in 12 medical datasets with an average accuracy of **82.01** percentages in comparison to 80.15 percentage of baseline Naïve Bayes algorithm.
- Average feature selection rate of the proposed model in the medical dataset appears 41.09 percentages.
- Average discretization rate of the proposed ensemble model appears to be 66.77 percentages.
- Highest accuracy is achieved in thyroid dataset.
- Highest accuracy improvement of 7.5 % is acquired in the Spectfheart dataset in comparison to baseline method.

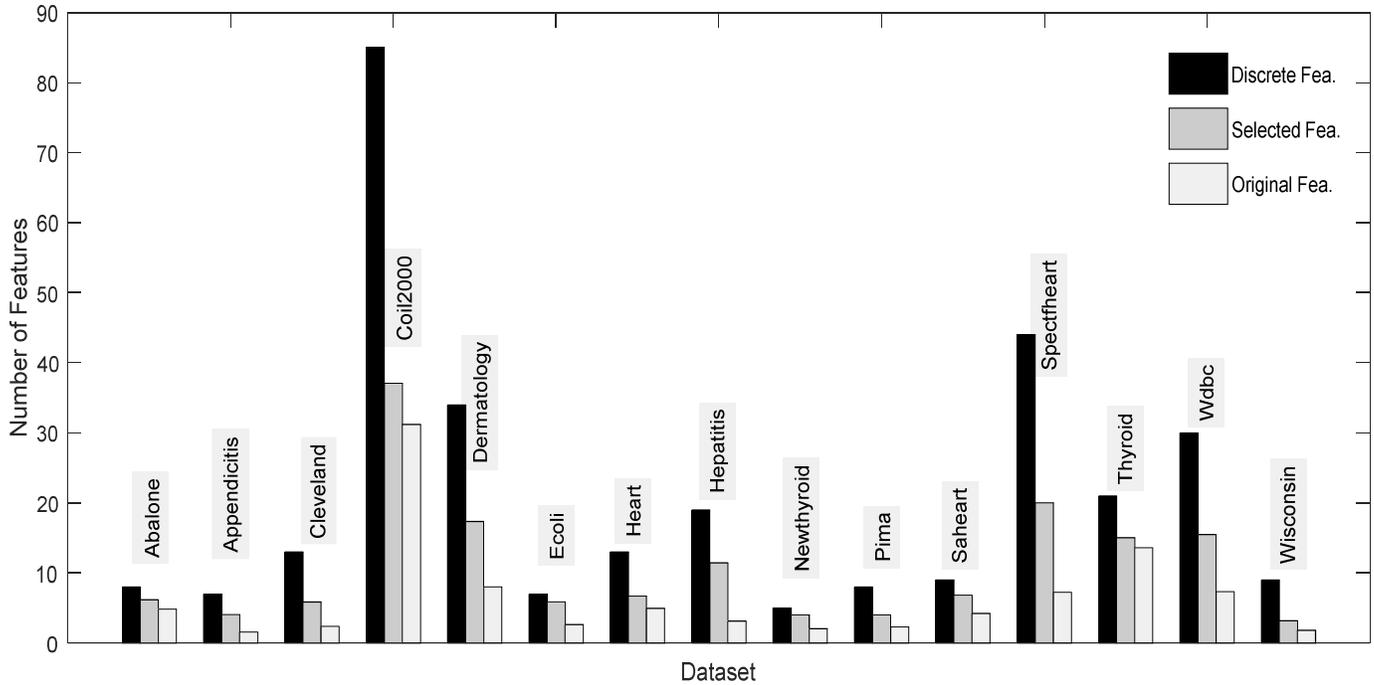

**Fig. 8: Attribute Reduction by the proposed model (Naïve Bayes)**

Fig. 8 shows the reduction rate of each dataset with the help of bar chart. This chart carries the three different color of bars where black bar represents the original number of features present in the dataset, dark grey color specifies the selected feature by the optimal solution, and light grey bar represents the number of discretized features.

**Table 4: Support Vector Machine performance and its comparison on Accuracy measure**

| Data Name | Features | Selected Features | Discretized | No. of fronts | Accuracy | Std. | Proposed Accuracy | Std. |
|---|---|---|---|---|---|---|---|---|
| Abalone | 8 | 6.51 | 5.25 | 14.65 | 0.254 | 0.020 | **0.264** | 0.019 |
| Appendicitis | 7 | 2.55 | 2.74 | 12.80 | 0.877 | 0.081 | **0.887** | 0.104 |
| Cleveland | 13 | 9.35 | 6.49 | 24.75 | 0.579 | 0.061 | **0.589** | 0.068 |
| Coil2000 | 85 | 34.29 | 27.30 | 28.38 | 0.914 | 0.004 | **0.948** | 0.002 |
| Dermatology | 34 | 20.25 | 7.11 | 26.53 | 0.975 | 0.031 | 0.969 | 0.031 |
| Ecoli | 7 | 5.72 | 2.04 | 32.65 | 0.777 | 0.078 | **0.815** | 0.047 |
| Heart | 13 | 7.32 | 6.45 | 16.17 | 0.837 | 0.050 | **0.844** | 0.065 |
| Hepatitis | 19 | 9.34 | 6.86 | 18.41 | 0.875 | 0.083 | **0.925** | 0.087 |
| Newthyroid | 5 | 2.84 | 1.15 | 21.45 | 0.958 | 0.056 | **0.968** | 0.048 |
| Pima | 8 | 4.88 | 2.74 | 32.08 | 0.767 | 0.039 | **0.779** | 0.023 |
| Saheart | 9 | 5.50 | 2.84 | 19.71 | 0.729 | 0.049 | 0.716 | 0.052 |
| Spectfheart | 44 | 26.04 | 21.75 | 27.24 | 0.794 | 0.032 | **0.820** | 0.062 |
| Thyroid | 21 | 13.66 | 9.19 | 16.25 | 0.996 | 0.003 | **0.996** | 0.003 |
| Wdbc | 30 | 22.41 | 8.76 | 28.62 | 0.972 | 0.015 | **0.984** | 0.015 |
| Wisconsin | 9 | 7.78 | 5.07 | 25.43 | 0.969 | 0.021 | **0.979** | 0.014 |

Table 4 collects the experimentation results of support vector machine classifier; we reported the SVM accuracy on original data as well as reduced data generated by the proposed model. On observing the empirical results, the following inference can be made:

- Proposed model outperform in 13 medical datasets with an average accuracy of **83.22** percentages in comparison to 80.15 percentage of without using proposed model.
- The average feature selection rate of the proposed model in the medical dataset seems to be 37.26 percentages.
- The average discretization rate of the proposed ensemble model appears to be 60.66 percentages.
- Highest accuracy is achieved in thyroid dataset.
- Highest accuracy improvement of 5 % is acquired in Hepatitis dataset in comparison to the baseline method

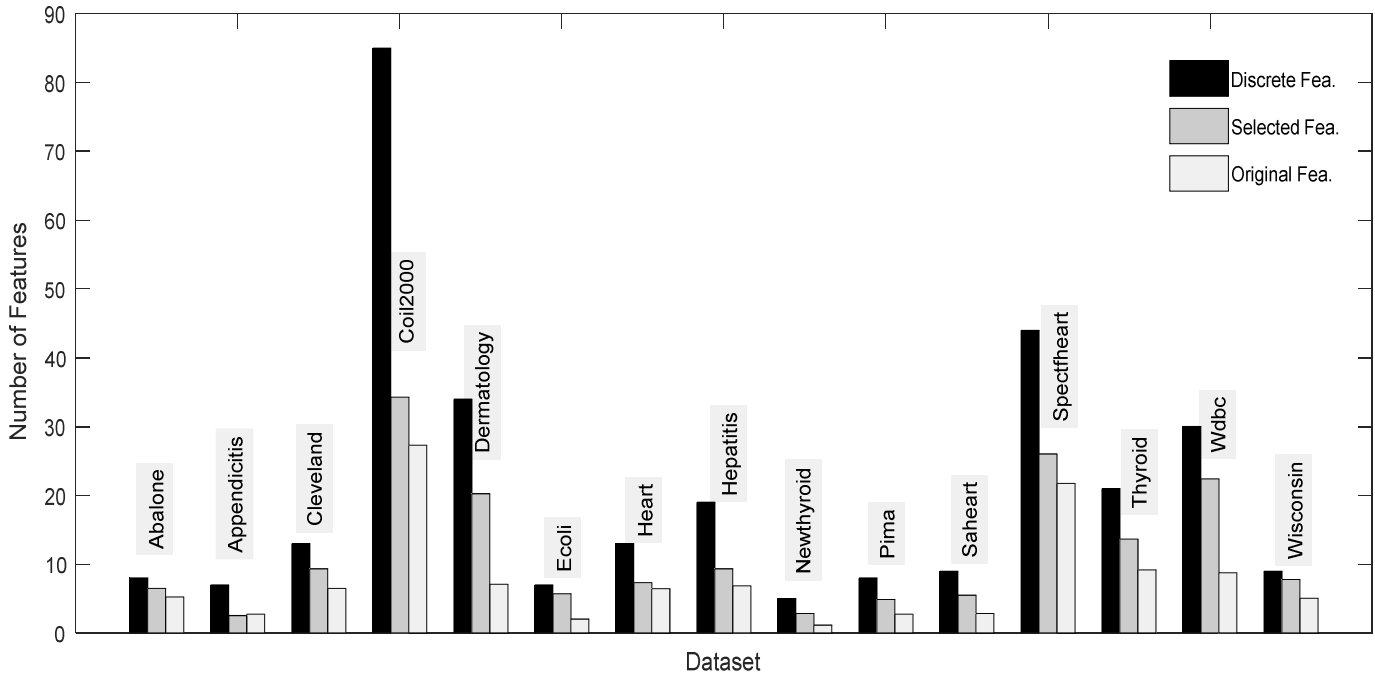

**Fig. 9: Attribute Reduction by the proposed model (SVM)**

The dimensionality reduction can be seen in Fig. 9, which indicates the effective reduction in the number of features achieved during feature selection and discretization phase when SVM is elected as the classifier.

**Table 5: Decision Tree (C4.5) performance and its comparison on Accuracy measure**

| Data Name | Features | Selected Features | Discretized | No. of fronts | Original Accuracy | Std. | Proposed Accuracy | Std. |
|---|---|---|---|---|---|---|---|---|
| Abalone | 8 | 7.92 | 6.40 | 26.89 | 0.217 | 0.015 | **0.252** | 0.022 |
| Appendicitis | 7 | 2.95 | 1.26 | 10.04 | 0.819 | 0.143 | 0.784 | 0.106 |
| Cleveland | 13 | 8.47 | 5.35 | 17.00 | 0.471 | 0.092 | **0.636** | 0.094 |
| Coil2000 | 85 | 30.22 | 29.31 | 28.23 | 0.914 | 0.004 | **0.940** | 0.001 |
| Dermatology | 34 | 33.07 | 32.25 | 12.18 | 0.950 | 0.032 | **0.967** | 0.026 |
| Ecoli | 7 | 6.87 | 5.53 | 14.13 | 0.788 | 0.045 | **0.833** | 0.058 |
| Heart | 13 | 8.43 | 6.92 | 11.92 | 0.752 | 0.109 | **0.852** | 0.063 |
| Hepatitis | 19 | 18.53 | 16.77 | 21.60 | 0.838 | 0.167 | **0.913** | 0.119 |
| Newthyroid | 5 | 3.99 | 1.91 | 20.74 | 0.940 | 0.032 | 0.939 | 0.039 |
| Pima | 8 | 6.66 | 4.85 | 28.20 | 0.723 | 0.045 | **0.749** | 0.045 |
| Saheart | 9 | 5.84 | 3.17 | 17.57 | 0.647 | 0.059 | **0.747** | 0.065 |
| Spectfheart | 44 | 37.21 | 32.46 | 11.09 | 0.697 | 0.074 | **0.782** | 0.096 |
| Thyroid | 21 | 17.69 | 12.00 | 25.82 | 0.996 | 0.003 | **0.996** | 0.003 |
| Wdbc | 30 | 24.47 | 22.92 | 15.74 | 0.919 | 0.033 | **0.968** | 0.025 |
| Wisconsin | 9 | 8.83 | 6.56 | 14.85 | 0.946 | 0.033 | **0.972** | 0.021 |

The summary of decision tree performance is illustrated in table 5 and Fig. 10. Table 5 assembles the accuracy values and the relevant improvement in the reduction rate on the standard biomedical dataset. The value in the table states the following facts:

- Proposed model outperform in 12 medical datasets with an average accuracy of **82.20** percentages in comparison to 80.15 percentage of without using proposed model.
- The average feature reduction rate of the proposed model in the medical dataset seems to be 21.60 percentages.
- The average discretization rate of the proposed ensemble model appears to be 39.78 percentages.
- Highest accuracy is achieved in thyroid dataset.
- Highest accuracy improvement of 16.5 % is acquired in Cleveland dataset in comparison to the baseline method

The feature reduction and discretized rate can be observed from the Fig. 10. This figure represents the amount of feature reduced by the feature selection and amount of feature to be considered during the discretization.

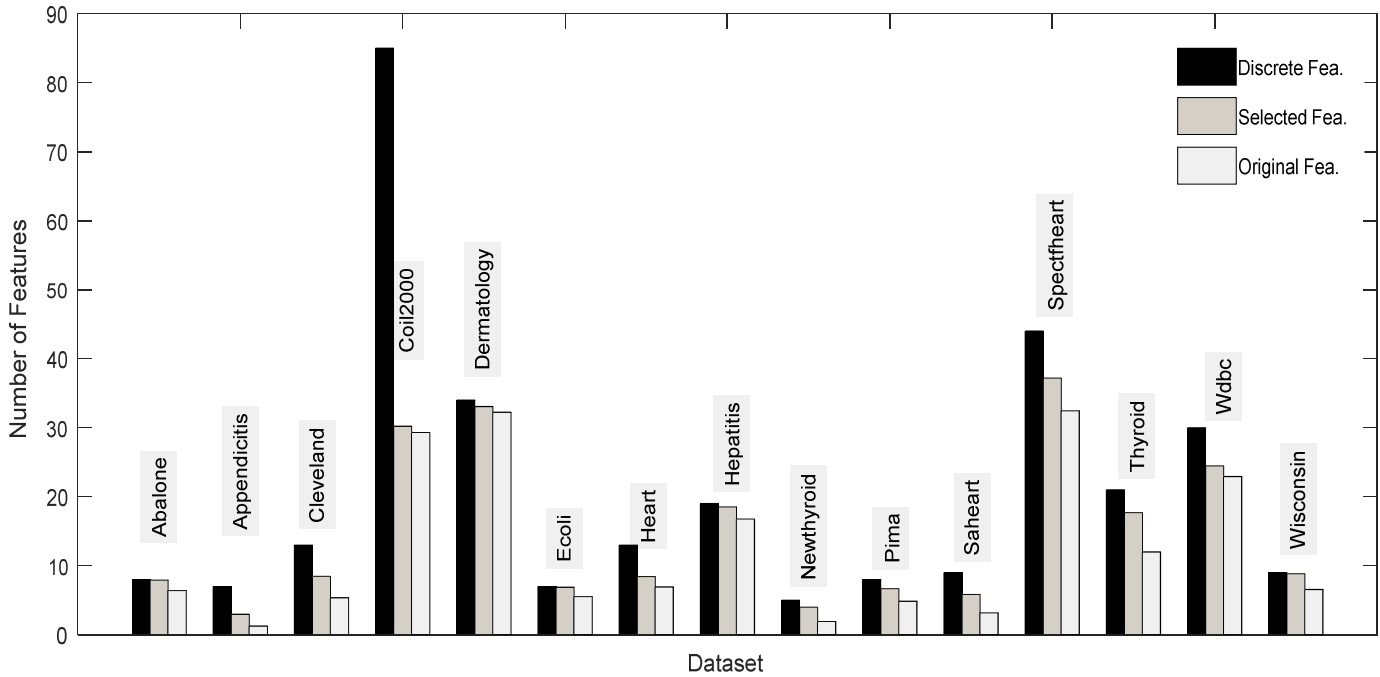

**Fig. 10: Attribute Reduction by the proposed model (Decision Tree)**

### 4.3. Multi-objective optimization

With multi-objective optimization, we end up needing the concept of dominance to claim the betterment of one solution over the other. The optimal solution of a multiple objective optimization problems is often known as the Pareto front a set of solutions, in contrast to single objective optimization where only single solution exists. To be precise, when coping with conflicting solutions some trade-offs will have got to be considered, and the most likely scenario is that you don't have a single solution for your issue, but a set regarding non-comparable solutions. One solution may be "better" regarding the particular objective but "worse" when it comes to one of many conflicting ones. The main goal will be then to find the particular solution which optimizes just about all the objectives at the same time.

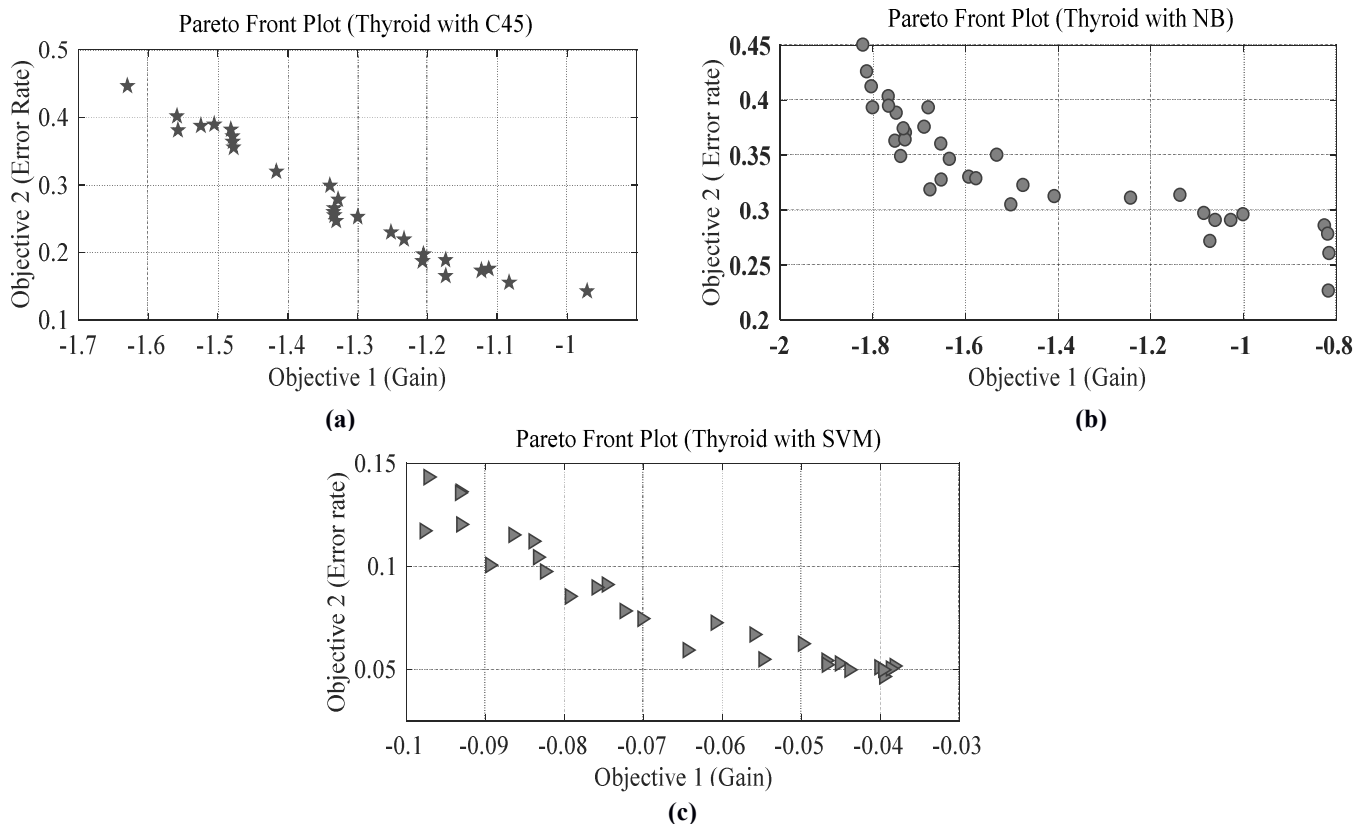

**Fig. 11. Pareto-optimal front produced by NSGA-II on Thyroid Dataset (a) Decision Tree (b) Naïve Bayes (c) SVM**

To visualize the multi-objective result, Fig. 11 and Fig. 12 presents the distribution of thyroid and coli datasets on the three base learners. The trade-off between the error rate and information gain is used here to evaluate the quality of solutions. Based on the Pareto chart, it is observed that SVM well approximate the Pareto-optimal front, depending on the size of the external non-dominated set. In comparison to C45 and Naïve Bayes, it evolved more Pareto-optimal solutions and distributed them more uniformly along the tradeoff front.

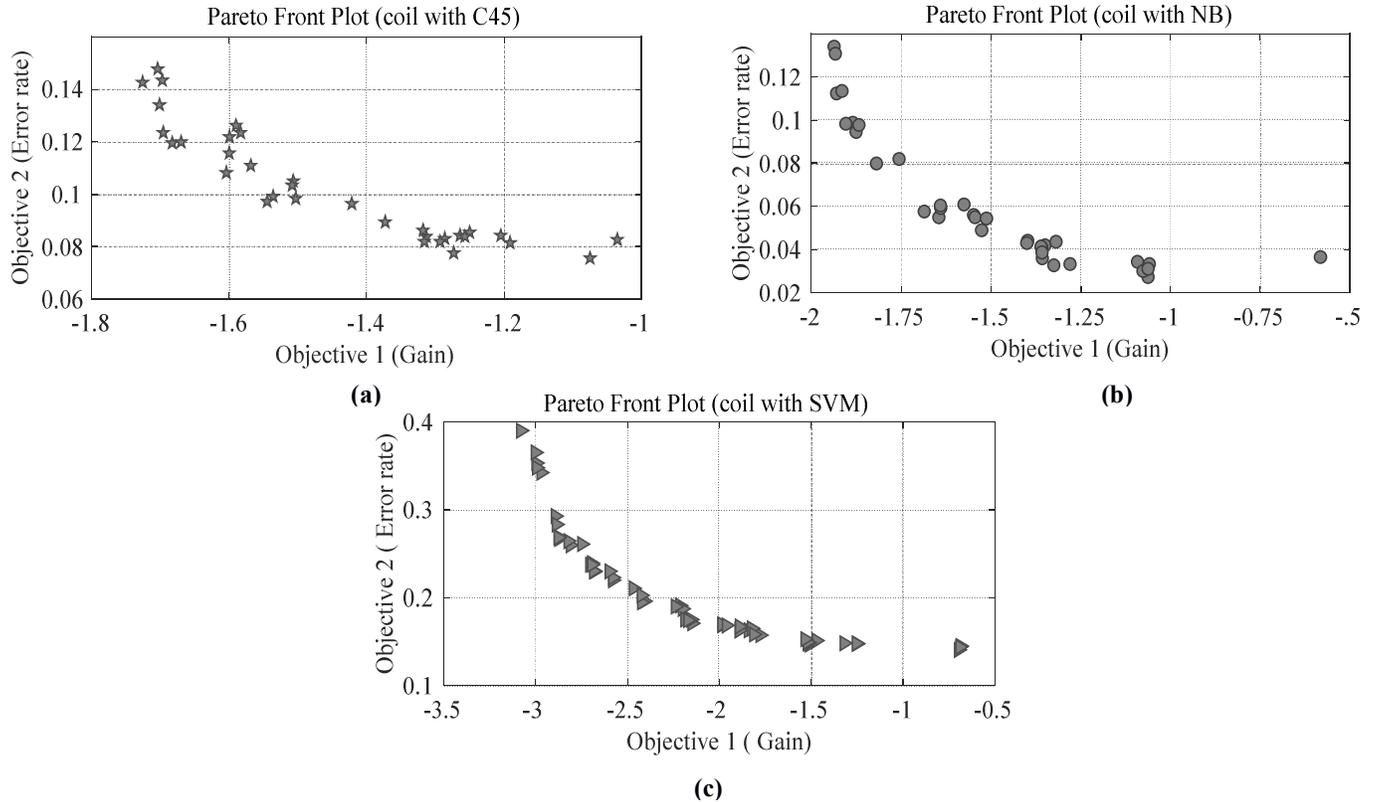

**Fig. 12. Pareto-optimal front produced by NSGA-II on Coil Dataset (a) Decision tree (b) Naïve Bayes (c) SVM**

## 5. Conclusion

This work presents an ensemble approach of solving the classification problem by considering feature selection and discretization both as a combined unit. The feature subset generation and the choice of discretized features are achieved through Multi-objective genetic algorithm NSGA-II. The subsets of the selected features are binary discretized for handling the wide range of continuous attributes. Feature subset is evaluated on the minimal error rate measure while the discretized features are evaluated on the maximum information gain. This proposed ensemble model has the advantage of dimension reduction in feature set and also across the data distribution of the feature set. Feature set reduction has been achieved by wrapper-based feature selection method and data distribution reduction is achieved by the binary discretization technique. Experimental result shows the effectiveness of proposed ensemble preprocessed model, by finding out the optimum number of features and decision of cut point in discretizing the features which results into the higher classification accuracy. The proposed method obtains the highest accuracy in more than twelve medical datasets and the possible reason being the diverse solution found by Pareto Optimal set. Moreover, the result indicates the proposed ensemble method has stronger performance aspect on support vector machine in classification. For future work, the splitting criteria of discretization could be a multi cut points that partition the data into multi interval and the use of other evolutionary algorithms with a high number of objectives function can be a promising research path.